\documentclass{article}

\usepackage{neurips_2019}

\usepackage[utf8]{inputenc} 
\usepackage[T1]{fontenc}    
\usepackage{hyperref}       
\usepackage{url}            
\usepackage{booktabs}       
\usepackage{amsfonts}       
\usepackage{nicefrac}       
\usepackage{microtype}      
\usepackage[pdftex]{graphicx}
\usepackage{algorithm}
\usepackage{algorithmic}

\title{Learning to Schedule DAG Tasks}
\author{%
  Zhigang Hua \\
  Ant Group\\
  Sunnyvale, CA 94085 \\
  \texttt{z.hua@antgroup.com} \\
  \And
  Feng Qi \\
  Ant Group\\
  Sunnyvale, CA 94085 \\
  \texttt{feng.qi@antgroup.com} \\
  \And
  Gan Liu \\
  Ant Group\\
  Hangzhou, China \\
  \texttt{liugan.lg@iantgroup.com} \\
  \And
  Shuang Yang \\
  Ant Group\\
  Sunnyvale, CA 94085 \\
  \texttt{shuang.yang@antgroup.com} \\
 }
 
\begin{document}
\setcitestyle{numbers}
\setcitestyle{square}
\maketitle

\begin{abstract}
Scheduling computational tasks represented by directed acyclic graphs (DAGs) is challenging because of its  complexity. Conventional scheduling algorithms rely heavily on simple heuristics such as \emph{shortest job first} (SJF) and \emph{critical path} (CP), and are often lacking in scheduling quality. In this paper, we present a novel learning-based approach to scheduling DAG tasks. The algorithm employs a reinforcement learning agent to iteratively add directed edges to the DAG, one at a time, to enforce ordering (i.e., priorities of execution and resource allocation) of ``tricky" job nodes. By doing so, the original DAG scheduling problem is dramatically reduced to a much simpler proxy problem, on which heuristic scheduling algorithms such as SJF and CP can be efficiently improved. Our approach can be easily applied to any existing heuristic scheduling algorithms. On the benchmark dataset of TPC-H, we show that our learning based approach can significantly improve over popular heuristic algorithms and consistently achieves the best performance among several methods under a variety of settings.
\end{abstract}



\maketitle

\section{Introduction}
Scheduling computational tasks, as commonly represented by directed acyclic graphs (DAGs), is an important problem in many areas of computer science ranging from programming language (e.g., compilation), operating systems (e.g., parallel processing), data engineering (e.g., distributed batch/streaming computation topology), to machine learning (e.g., training graphs). Exact algorithms are prohibited due to the exponential complexity -- the problem is known to be NP-hard for most of its variants. Current cluster schedulers rely entirely on simple heuristics such as \emph{shortest-job-first} (SJF), \emph{critical path} (CP), \emph{first-in-first-out (FIFO)}, and \emph{Tetris} \cite{grandl2014multi}\cite{mao2019learning}\cite{cham2010}\cite{isard2007}\cite{zaha2012}. These heuristics are problem independent, incapable to utilize the dependencies as defined by the DAGs or the resource consumption constraints when scheduling a job. As a result, these methods are often far from optimal. 

In this paper, we present a deep reinforcement learning algorithm for DAG scheduling. Our approach is motivated by the observation that unsatisfactory schedules are very often due entirely to the wrong ordering of a few job nodes while the rest is close to optimal. If these ``tricky" nodes are correctly ordered, the quality of schedules would be dramatically improved. Moreover, if the correct ordering of these nodes can be given a priori by an oracle, the scheduling tasks would become substantially simpler such that even very simple heuristics are able to find near-optimal solutions. One way to achieve this is to add directed edges to break the ties among tricky job nodes, i.e., to explicitly require that one job node needs to be given priority over another when it comes to execution as well as resource allocation. For example, as shown in Figure \ref{example}, after adding directed edges, both SJF and CP are able to find optimal schedules. By adding directed edges, we simplify DAG scheduling by including additional constraints and convert the original problem to a simpler proxy. In principle, this is similar in spirit to classic OR approaches such as \emph{cutting plane} or \emph{active set}.

We present a neural architecture that learns to add directed edges to a given DAG iteratively over time. The proposed model includes a GNN component \cite{lecun2015} that is used to effectively
encode the problem instance including the topological structure of the DAG as well as the problem configurations such as the runtime and resource requirements of each job node. On top of it, a policy network is designed to select potential directed edges that can be added to the DAG. The network is trained end-to-end using reinforcement learning to minimize the makespan. Our approach can be easily integrated with any existing heuristic algorithms without modification, e.g., SJF, CP, and so on. We test our method on public benchmark TPC-H. Experiment results show that it can effectively improve heuristic algorithms for DAG scheduling. 

\section{The DAG Scheduling Problem}
The DAG scheduling problem is a combination of two well-known NP-hard problems: the minimum makespan problem \cite{vazirani2013approximation} and the bin packing problem \cite{hopper2001review}\cite{martello2000three}.  The goal is to find an optimal scheduling order that is executed with minimal makespan. The former features interdependence among tasks, and the latter features one-dimensional or multi-dimensional resource constraints.  The Minimum Makespan Scheduling Problem aims at allocating dependent job nodes on limited computation resources with the goal of minimizing the finishing time, aka makespan. The solution can be found by solving an integer programming problem with branch-and-bound \cite{clausen1999branch}\cite{kohler1974characterization}, which is unfortunately intractable for most practical-scale problems. Heuristic approaches such as highest level first, longest job time, critical path, and random priority \cite{kwok1999static}, assign priorities to tasks and then execute tasks when their dependent tasks are finished, which are fast but provide sub-optimal solutions. In a dynamic makespan problem, the DAG jobs are not known beforehand, but arrive dynamically in a stream \cite{palis1995online}\cite{mao2019learning}. In a preemptive makespan problem, any job execution can be paused or restarted \cite{chen1988preemptive}. In a probabilistic or conditional makespan problem, task branches are executed with uncertainty \cite{elrewini1995static}.

There has been a surge of interests recently in finding adaptive data-driven scheduling algorithms using machine learning. For example, \cite{deeprm} and \cite{chen2019} used machine learning to learn near-optimal ordering of job nodes. These studies did not consider the dependencies among the jobs as defined by the DAGs, neither did they exploit the DAG topology when allocating executors and other resources. \emph{Decima} \cite{mao2019learning} uses reinforcement learning to select the ordering of DAG jobs as well as the allocation of executors.

\begin{figure}
\centering
    \qquad
    \includegraphics[width=0.8\linewidth]{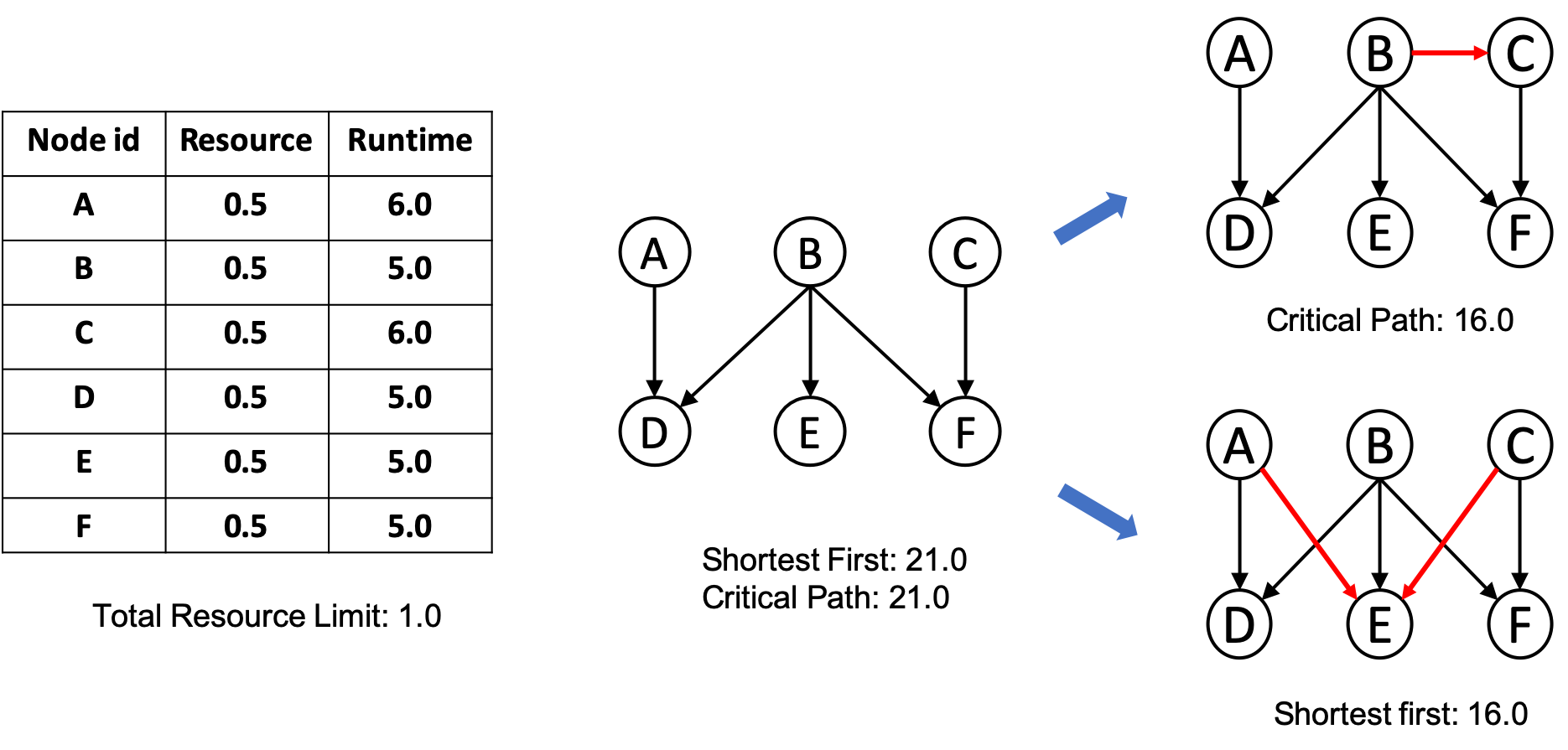}
\caption{An example of improving DAG scheduling by adding edges. Left: node resource and runtime of a DAG. Middle: DAG's makespan is 21.0 for both critical path (CP) and shortest job first (SJF), the CP algorithm ranks nodes by the maximum sum of task durations along the path to any of their leaf nodes in a DAG. Right: adding edges (red) helps reduce makespan of CP and SJF to 16.0.}
\label{example}
\end{figure}

We examine DAG scheduling in a more general form. The problem considered in this paper is illustrated in Figure \ref{example}, where each DAG's job node has two values, aka runtime and resource requirement (which is normalized by the total resource as between 0 and 1), and the total available resource is at a unit capacity of 1.0. Given a bunch of DAG jobs under a limited resource, the goal is to minimize the makespan of all DAG while no preemption of any DAG or node is allowed.

\section{Learning Algorithm} \label{la}
In the following subsections, we first introduce the reinforcement learning formulation, then the GNN models for graph representations of DAGs, the neural policy networks, and the training and inference algorithms.

\subsection{\textbf{Reinforcement Learning Formulation}\label{rlf}}
\begin{figure}[H]
\centering
\qquad
\includegraphics[width=1.0\linewidth]{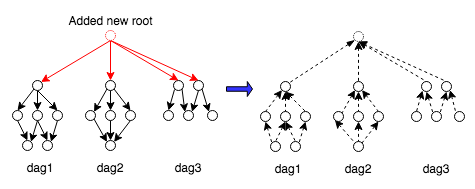}
\caption{Merge DAGs by adding a new root node that connects to all existing root nodes of DAGs (left), and reverse direction of edges when computing graph embedding for the merged DAG (right).}
\label{gnn}
\end{figure}

Given multiple job DAGs, we merge them into one graph $G$ by introducing a new root node that connects to existing root nodes of DAGs, as shown in Figure \ref{gnn} left. We employ deep reinforcement learning algorithm that learns to add edges in $G$ for better scheduling. The addition of directed edges in a DAG can be formulated as a Markov Decision Process (MDP) with following components:

\begin{itemize}
\item \textbf{State}: a DAG graph $G=\{n, N^{(0)}, E^{(0)}\}$, where $n$ is the number of nodes, $N^{(0)}$ is the set of $n$ nodes, and $E^{(0)}$ is the set of directed edges.
\item \textbf{Action}: adding a directed edge $a$ in $G$ by sequentially selecting a starting node $a^1$ and an ending node $a^2$ (see Figure \ref{example} right). Here an added edge should not conflict with graph structure of $G$, and a conflict occurs when there already exists a directed path between $a^1$ and $a^2$ in $G$. The selection of starting node $a^1$ is conditioned on $G$, and the selection of ending node $a^2$ is conditioned on $G$ and $a^1$. By sequentially selecting the starting and ending nodes, the time complexity of adding an edge in $G$ is reduced from $O(n^2)$ to $O(n)$.
\item \textbf{Transition}: $G'$ is composed by adding an edge $a$ in $G$.
\item \textbf{Reward}: in our experiments, we take reward as the difference between makespans of $G'$ and $G$ using existing heuristic methods such as SJF, CP, etc. The reward can be easily switched to other measures than makespan, such as average waiting time.
\end{itemize}

\subsection{\textbf{Graph Neural Network}\label{grod}}
We employ GNN to compute graph embeddings for DAGs. Given a DAG $G$, we reverse the direction of all edges in $G$ to get $G^T$. 
The original feature vector of each node in a DAG is comprised of its runtime and required resources, which is transformed into a hidden representation $e^{(0)}$ through a $L$-layer neural network with parameters of $W_1^{(l)}$ ($1 \leq l \leq L$). 

At each iteration of message passing in $G^T$, each node's embedding is propagated to its neighbors and updated as follows:
\begin{equation} \label{eq1}
e_i^{(h)} = \sigma(W_{2}^{(h)} \cdot Concat( e_i^{(h-1)}, \sum_{j\in{N_i}} e_j^{(h-1)} )), \forall i\in\{1,...,n\}, \forall h\in\{1,...,H\}
\end{equation}

where $\sigma(\cdot)$ stands for activation function, $H$ is the number of message passing iterations and $h$ is the order, $W_2^{(h)}$ is the weight matrix for the $h$-th iteration, $N_i$ represents the set of neighbors in $G^T$ (aka children nodes in $G$). Hence the parameters of the GNN network are:
\begin{equation}\label{eq2}
\phi = \{W_1^{(l)}, W_2^{(h)}\}, \forall l\in\{1,...,L\}, \forall h\in\{1,...,H\}
\end{equation}

As the output the GNN model, each node is represented as a deep embedding, and the whole DAG graph is represented as an average embedding of all its included nodes.

\begin{figure}
\centering
\includegraphics[width=0.8\linewidth]{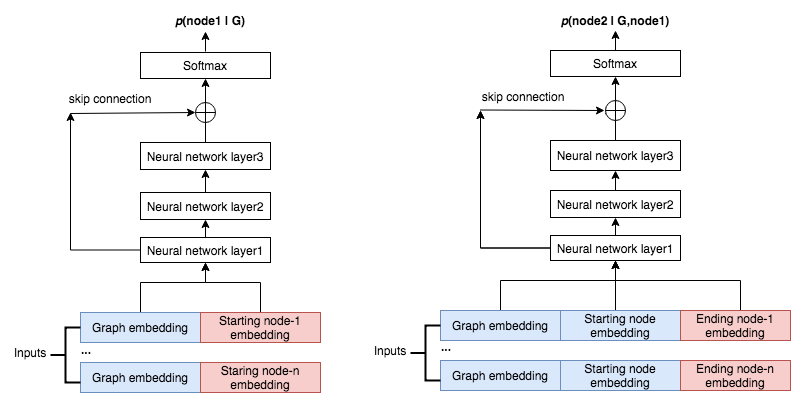}
\caption{Two policy networks for selecting starting (left) and ending nodes (right) when adding an edge in a DAG. For inputs, blue boxes are identical and red boxes are different for candidate nodes.}
\label{policy_network}
\end{figure}

\subsection{\textbf{Neural Policy Network}} \label{nn}
Our method takes an action of adding an edge $a$ for the DAG graph $G$, which sequentially selects starting node $a^1$ and ending node $a^2$. The joint probability of selecting $a$ is the multiplying product of the probabilities of selecting $a^1$ and $a^2$ as follows:

\begin{equation}\label{eq3}
\pi \sim p_{\theta,\phi} (a|G) = p_{\theta1,\phi}(a^1|G) \cdot p_{\theta2,\phi}(a^2|G, a^1)
\end{equation}

where $\theta=(\theta_1, \theta_2)$, $p_{\theta1,\phi}(a^1|G)$ is the conditional probability of selecting starting node given $G$, and $p_{\theta2,\phi}(a^2|G, a^1)$ is the conditional probability of selecting ending node given $G$ and pre-selected $a^1$. We use two deep neural networks to compute these conditional probabilities.

Due to its compelling results we adopt Reset \cite{he2016} to represent two deep policy networks, as shown in Figure \ref{policy_network} that selects starting (left) and ending node (right) for adding an edge. The neural network's input is encoded as a combination of multiple embeddings, including the graph, the starting node if selecting an ending node, and candidate nodes. As output of the policy networks, the conflicting nodes are masked with probability of 0 at the last $softmax$ layer. It is crucial for a learning method that trains on given graphs and predicts on unseen graphs. This neural architecture is generalized such that it supports heterogeneous DAG structures with different amount of nodes.



\subsection{\textbf{Training Algorithm}} \label{trainingalg}
The GNN and policy networks are imporved based on policy gradient algorithm, which is summarized in Appendix \ref{trainalgo}. We discuss its main components as follows:

\textbf{Policy rollout}. Given a DAG graph $G$, we do $N$ rollouts of current policy models and in each rollout we add $M$ directed edges. Based on a commonly used practice of encouraging exploration, we use $\epsilon$-greedy \cite{sutton2018} such that the rollout will pick randomly each node with a probability of 0.05 at every timestep. The rollout algorithm is presented in Appendix \ref{rollout}. At $j$-th timestep of $i$-th rollout, we record a triplet to be used by our training algorithm as follows:

\begin{equation}\label{triplet}
\{G_{i,j}, (a^1_{i,j}, a^2_{i,j}), r_{i,j}\}, \forall i\in\{1,...,N\}, \forall j\in\{1,...,M\} 
\end{equation}

where $G_{i,j}$ is the updated DAG graph with added edges, $a^1_{i,j}$ and $a^2_{i,j}$ stands for the starting node and ending node of an added edge $a_{i,j}$ respectively, and $r_{i,j}$ is the reward computed as the reduction of makespan of $G_{i,j}$ after an edge $a_{i,j}$ is added.

\textbf{Reward}. At each timestep $j$, the incremental reward is the reduction of makespan of a DAG after an edge is added. To inspire long term reward, the cumulative reward at the $j$-th timestamp is a discounted sum of incremental rewards from current timestep until completion of a rollout:

\begin{equation}\label{reward}
\widehat r_{i,j} = \sum_{j'=j}^M r_{i,j'} \cdot \gamma^{(j'-j)}, \forall i\in\{1,...,N\}, \forall j\in\{1,...,M\} 
\end{equation}
where $\gamma$ is the discounting factor, and by default we set it as 1.0 in our experiments.

\textbf{Baseline}. For every $N$ rollouts, we compute the baseline of rewards at each timestep $j$ as shown in Equation (\ref{base}), which is subtracted from rewards to reduce variance.
\begin{equation}\label{base}
b_j = 1/N \cdot \sum_{i=1}^N \widehat r_{i,j}, \forall j\in\{1,...,M\} 
\end{equation}

\textbf{Loss function}. The model parameters in the reinforcement learning algorithm are updated after every $N$ rollouts. We use the commonly used policy gradient algorithm to train the parameters in the GNN and policy networks. The joint loss function is represented as follows:
\begin{equation}\label{loss1}
\nabla _{\phi, \theta} J(\phi, \theta|G) = \mathop{\mathbb{E}} _{\pi\sim{p_{\phi, \theta}(.|G)}} (\widehat r - b) \cdot \nabla \log p_{\phi, \theta}(\pi|G)
\end{equation}

With $N$ rollouts where $M$ edges are added, Equation (\ref{loss1}) can be  computed as follows:
\begin{equation}\label{loss2}
\nabla 
J(\phi,\theta_1,\theta_2|G) = \sum_{i=1}^N\sum_{j=1}^M (\widehat r_{i,j} - b_j) \cdot [\nabla \log p_{\phi,\theta_1}(a_{i,j}^1|G_{i,j}) + \nabla \log p_{\phi,\theta_2}(a_{i,j}^2|G_{i,j},a_{i,j}^1)]
\end{equation}



\subsection{\textbf{Inference Algorithm}} \label{infalg}
We use a beam search method to add edges for DAGs in the inference stage. Suppose the beam size is $K$ (e.g. $K=10$). The inference algorithm is shown in Appendix \ref{inf_alg}. When adding an edge, we first pick up the top $K$ candidate starting nodes based on their probabilities output by the policy network, and then for each of $K$ starting nodes we choose its top ending node with maximal probability. As a result, the edge is selected with the maximal joint probability of the $K$ candidates.

\begin{figure}
\centering
\qquad
\includegraphics[width=1.0\linewidth]{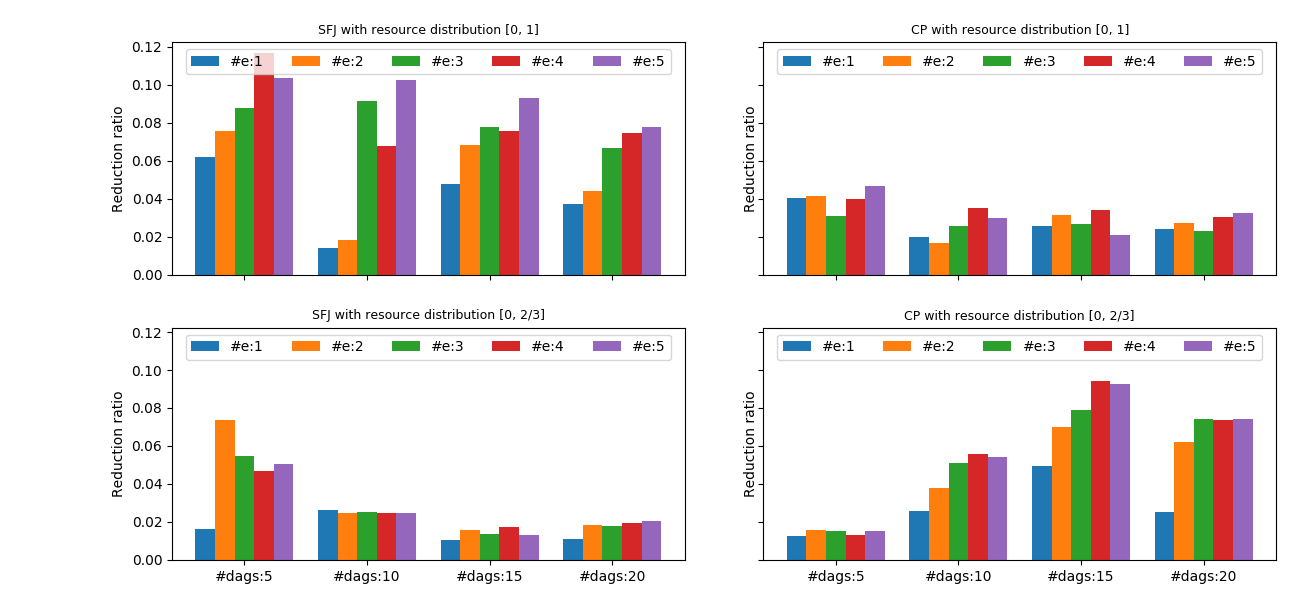}
\caption{The reduction ratios of DAG makespans for SFJ and CP with learning method by varying the number of added edges (from 1 to 5) under two job resource distributions.}
\label{edges}
\end{figure}

We applied our learning method to two heuristic algorithms of SJF and CP. We tested Tetris \cite{grandl2014multi} and mixed integer linear programming (MILP) for comparison. The CP algorithm ranks nodes by the maximum sum of task durations along the path to any of their connected leaf nodes. The Tetris algorithm ranks nodes by the dot product similarity of two vectors of resource consumption and available resource. The MILP formulation is described in Appendix \ref{milp_form}. As its scale is large for existing IP solvers, we relax the integer constraints to solve the linear programming problem (LP) with OR-Tools \footnote{https://developers.google.com/optimization}, which might produce sub-optimal results. 

\begin{table}[htbp]
\caption{The average makespan (sec) with real-valued runtime and resource dist of 1.0.}
\label{r1}
\centering
\begin{tabular}{c|ccc|ccc|c|c}\hline
\multicolumn{1}{c|}{} & 
\multicolumn{3}{c|}{Shortest Job First (SJF)} & 
\multicolumn{3}{c|}{Critical Path (CP)} &
\multicolumn{1}{c|}{OR-Tools} &
\multicolumn{1}{c}{Tetris}
\\ \hline
\# dags & Time & Learn & Reduce \% & Time & Learn & Reduce \% & Time & Time \\ \hline
5 & 21.29 & 18.81 & 11.65\% & 18.13 & \textbf{17.29} & 4.66\% & 23.05 & 17.73\\ 
10 & 58.66 & 52.66 & 10.23\% & 46.66 & \textbf{45.03} & 3.50\% & 59.56 & \textbf{45.03}\\ 
20 & 87.66 & 80.85 & 7.77\% & 72.35 & \textbf{70.01} & 3.23\% & 97.09 & 70.94 \\  
50 & 241.33 & 231.80 &3.95\% & 203.72 & 191.67 & 5.92\% & - & \textbf{191.47} \\
100 & 479.54 & 478.19 &0.28\% & 404.22 & 384.74 & 4.82\% & - &\textbf{382.61} \\\hline
average 	& 177.70	& 172.46 &	6.78\%	 & 149.02 	& 141.75 	& 4.43\%	&-& \textbf{141.56}  \\ \hline 

\end{tabular}
\vspace{0.25cm}


\caption{The average makespan (sec) with real-valued runtime and resource dist of 2/3.}
\label{r0.67}
\centering
\begin{tabular}{c|ccc|ccc|c|c}\hline
\multicolumn{1}{c|}{} & 
\multicolumn{3}{c|}{Shortest Job First (SJF)} & 
\multicolumn{3}{c|}{Critical Path (CP)} &
\multicolumn{1}{c|}{OR-Tools} &
\multicolumn{1}{c}{Tetris}
\\ \hline
\# dags & Time & Learn & Reduce \% & Time & Learn & Reduce \% & Time & Time \\ \hline
5 & 14.55 & 13.48 & 7.36\% & 11.94 & \textbf{11.73} & 1.76\% & 13.49 & 13.26\\ 
10 & 38.45 & 37.15 & 3.37\% & 33.40 & \textbf{31.54} & 5.56\% & 37.69 & 33.03 \\ 
20 & 61.04 & 59.79 & 2.05\% & 53.71 & \textbf{49.72} & 7.44\% & 59.64 & 50.93 \\ 
50 & 118.54	& 111.83 &	5.66\% &	 98.44 &	  \textbf{97.94} &	0.50\%	& - & 99.46 \\
100	& 229.05 &	220.81 &	3.60\%	& 189.59 &	  \textbf{188.94} &	0.35\%	& - & 189.88 \\ \hline
average 	& 92.33	& 88.61 &	4.41\%	 & 77.42 	& \textbf{75.97}	& 3.12\%	&-& 77.31  \\ \hline
\end{tabular}
\vspace{0.25cm}

\caption{The average makespan (sec) with real-valued runtime and resource dist of 1/3.}
\label{r0.33}
\centering
\begin{tabular}{c|ccc|ccc|c|c}\hline
\multicolumn{1}{c|}{} & 
\multicolumn{3}{c|}{Shortest Job First (SJF)} & 
\multicolumn{3}{c|}{Critical Path (CP)} &
\multicolumn{1}{c|}{OR-Tools} &
\multicolumn{1}{c}{Tetris}
\\ \hline
\# dags & Time & Learn & Reduce \% & Time & Learn & Reduce \% & Time & Time \\ \hline
5 & 8.49 & 7.84 & 7.60\% & 7.11 & \textbf{7.02} & 1.29\% & 7.70 & 8.77\\ 
10 & 19.39 & 18.59 & 4.15\% & 16.10 & \textbf{15.56} & 3.36\% & 17.25 & 16.81 \\ 
20 & 29.49 & 28.02 & 4.99\% & 24.06 & \textbf{23.47} & 2.44\% & 27.01 & 25.02 \\
50 &	68.73 &	66.52 &	3.22\%	& 60.02 &	  \textbf{59.31} &	1.19\%	& -	 &60.85 \\
100 & 138.21 &	136.93 &	0.93\% & 123.13 &	  \textbf{122.31} &	0.66\%	& -	 &122.83 \\ \hline
average 	& 52.86	& 51.58 &	4.18\%	 & 46.08 	& \textbf{45.53}	& 1.79\%	&-& 46.86  \\\hline 
\end{tabular}
\end{table}

\section{Experiments and Results} \label{ExperimentsSection}
The TPC-H\footnote{http://www.tpc.org/tpch/} is a decision support benchmark that have broad industry-wide relevance. In its DAG datasets, the number of nodes vary from 2 to 18. In the training stage, at each iteration we sample a sequence of DAGs and merge them into one for training. For testing, we built four testsets to measure performance with number of DAGs at 5, 10, 15 and 20. For each testset we sample 10 sequences of DAGs and each sequence is merged into one DAG. The job resource distribution is specified as the ratio of the maximum node resource consumption over total resource, which reflects tightness of resource constraints, and is tuned to 3 values (i.e. 1.0, 2/3 and 1/3) in our experiments. The higher values mean tighter resource constraints.

\subsection{Effects of Added Edges}
As show in Figure \ref{edges}, we test our learning method by varying the number of added edges from 1 to 5. The experiment is conducted under two different resource distributions, e.g. 1.0 and 2/3. The results show that our leaning method can reduce makespan of DAG execution on both SJF and CP algorithms. The two figures in the left side show the reduction ratio of makespan using SJF, where the reduction ratio of makespan is up to about 12\%. The two figures in the right side show the reduction ratios of makespan using CP, where the reduction of makespan is up to about 10\%.  More added edges leads to better reduction of makespan but not always. In the following experiments, we use an ensemble way that chooses the best performance among 1 to 5 added edges in DAGs to measure makespan reduction under different distributions of resource and runtime values.

\begin{table}
\caption{The average makespan (sec) with uniformly random runtime and resource dist of 1.0.}
\label{r1_1}
\centering
\begin{tabular}{c|ccc|ccc|c|c}\hline
\multicolumn{1}{c|}{} & 
\multicolumn{3}{c|}{Shortest Job First (SJF)} & 
\multicolumn{3}{c|}{Critical Path (CP)} &
\multicolumn{1}{c|}{OR-Tools} &
\multicolumn{1}{c}{Tetris}
\\ \hline
\# dags & Time & Learn & Reduce \% & Time & Learn & Reduce \% & Time & Time \\ \hline
5 & 9.54 & 8.29 & 13.12\% & 8.95 & 7.91 & 11.61\% & 9.21 & \textbf{7.71} \\ 
10 & 20.82 & 19.70 & 5.37\% & 19.67 & 16.76 & 14.77\% & 20.24 & \textbf{16.26} \\ 
20 & 34.31 & 31.79 & 7.35\% & 31.03 & 28.22 & 9.05\%  & 35.29 & \textbf{26.92} \\
50	& 89.93	& 87.51 &	2.69\%	 & 83.49 	&  73.68 	& 11.76\%	&-&  \textbf{69.21}  \\
100	& 178.01	& 176.43 &	0.89\%	 & 173.88 	& 148.40 	& 14.66\%	&-&  \textbf{138.60}  \\ \hline
average 	& 66.52	& 64.74 &	5.88\%	 & 63.41 	& 54.99 	& 12.37\%	&-& \textbf{51.74}  \\\hline
\end{tabular}
\vspace{0.25cm}


\caption{The average makespan (sec) with uniformly random runtime and resource dist of 2/3.}
\label{r1_0.67}
\centering
\begin{tabular}{c|ccc|ccc|c|c}\hline
\multicolumn{1}{c|}{} & 
\multicolumn{3}{c|}{Shortest Job First (SJF)} & 
\multicolumn{3}{c|}{Critical Path (CP)} &
\multicolumn{1}{c|}{OR-Tools} &
\multicolumn{1}{c}{Tetris}
\\ \hline
\# dags & Time & Learn & Reduce \% & Time & Learn & Reduce \% & Time & Time \\ \hline
5 & 6.51 & 6.07 & 6.72\% & 5.96 & \textbf{5.39} & 9.49\%  & 6.53 & 6.24 \\ 
10 & 14.10 & 13.48 & 4.42\% & 12.48 & \textbf{11.77} & 5.70\% & 13.00 & 12.04  \\ 
20 & 22.73 & 21.79 & 4.12\% & 20.18 & \textbf{19.53} & 3.21\% & 21.40 & 20.25 \\  
50	& 39.56	& 39.10 &	1.17\%	 & 35.65 	&  \textbf{35.18 }	& 1.33\%	&-& 35.41  \\
100	& 79.18	& 78.47 &	0.89\%	 & 71.63 	&  70.98 	& 0.91\%	&-&  \textbf{70.72}  \\ \hline
average 	& 32.42	& 31.78 &	3.46\%	 & 29.18 	& 28.57 	& 4.13\%	&-& \textbf{28.93}  \\ \hline
\end{tabular}
\vspace{0.25cm}

\caption{The average makespan (sec) with uniformly random runtime and resource dist of 1/3.}
\label{r1_0.33}
\centering
\begin{tabular}{c|ccc|ccc|c|c}\hline
\multicolumn{1}{c|}{} & 
\multicolumn{3}{c|}{Shortest Job First (SJF)} & 
\multicolumn{3}{c|}{Critical Path (CP)} &
\multicolumn{1}{c|}{OR-Tools} &
\multicolumn{1}{c}{Tetris}
\\ \hline
\# dags & Time & Learn & Reduce \% & Time & Learn & Reduce \% & Time & Time \\ \hline
5 & 4.88 & \textbf{4.82} & 1.14\% & \textbf{4.82} & \textbf{4.82} & 0.00\% & 4.95 & 5.25 \\ 
10 & 7.67 & 7.07 & 7.82\% & 6.15 & \textbf{5.90} & 4.01\% & 6.95 & 6.92 \\ 
20 & 11.08 & 10.69 & 3.49\% & 9.63 & \textbf{9.28} & 3.67\% & 10.31 & 11.10 \\
50	& 25.11	& 24.85 &	1.03\%	 & 23.05 	&  \textbf{22.59} 	& 2.01\%	&-& 24.07  \\
100	& 51.54	& 51.23 &	0.60\%	 & 47.17 	&  \textbf{46.66} 	& 1.08\%	&-& 48.32  \\ \hline
average 	& 20.06	& 19.73 &	2.82\%	 & 18.16 	& \textbf{17.85} 	& 2.15\%	&-& 19.13  \\ \hline
\end{tabular}
\end{table}

\subsection{Resource Distribution} \label{ResDist}
We test our learning algorithm to measure its generalization and robustness under different job resource distributions. We test 4 different resource distributions. Table \ref{r1}-\ref{r0.33} list the comparison of SJF and CP under different resource distributions. Generally, under tighter resource constraints, there are more improvement spaces for learning algorithm to reduce makespans of SJF and CP. Overall the learning algorithm brings good improvements for both of the SJF and CP methods. Under different resource distributions, on average it achieves about 6.03\% and 4.30\% on the reduction of makespans for SJF and CP, respectively. Originally, CP is more efficient than SJF for DAG scheduling in our experiment, and the overall improvement of makespan for CP (4.30\%) is less than SJF (6.03\%). In this table, compared to other methods like SJF, Tetris and Or-tools, the learning-based algorithm for CP has consistently achieved the best results under different settings.

\subsection{Runtime Distribution} \label{RtDist}
In the above experiments, the runtime of job nodes in DAGs are set as real values in the TPC-H dataset. To test the robustness of our learning algorithm, we further tune the runtime values of DAG nodes using uniform distribution between 0 and 1.0. Table \ref{r1_1}-\ref{r1_0.33} show the results of our experiments. Under different settings, on average the learning algorithm can achieve reduction of makespans  of 6.0\% for SJF and 6.63\% for CP, respectively. As shown in the tables, the improvements of CP are more obvious for random runtime (6.63\%) than real-valued distribution (4.3\%), while for SJF the improvements are less significantly different between uniform runtime (6.0\%) and real values (6.03\%). The CP algorithm prioritizes the nodes in the critical path, and setting uniform random runtime for DAG nodes offers improvements for these nodes. The learning-based algorithm for CP achieves the overall best performance under all the settings, and Tetris claims the best result under the job resource distribution of 1.

\subsection{Convergence Analysis} \label{CA}
\begin{figure}
\centering
\qquad
\includegraphics[width=1\linewidth]{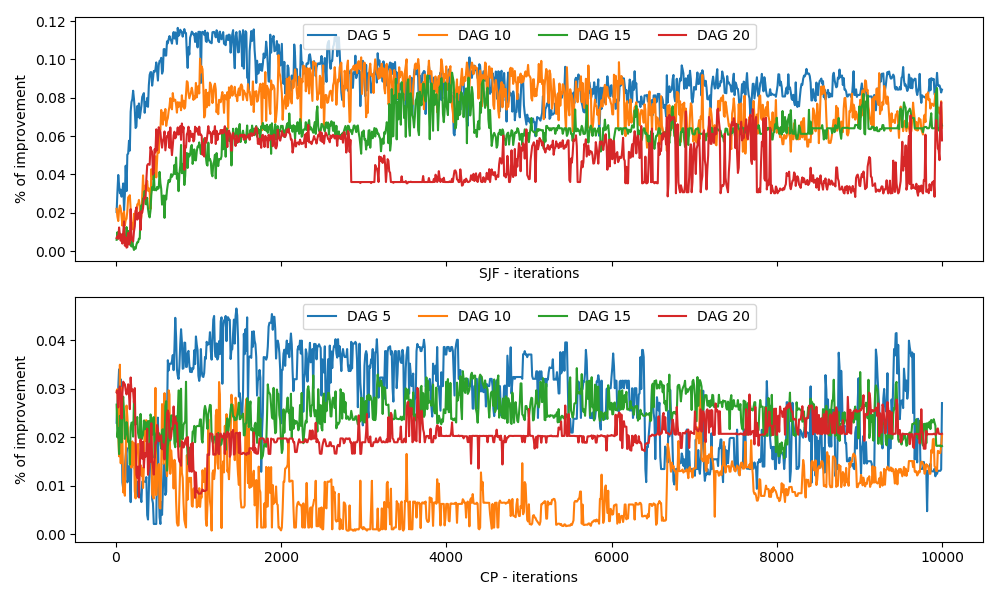}
\caption{Convergence of test datasets under job resource distribution of 1.0}
\label{convergence_1}

\end{figure}

We study the convergence of our learning algorithm on the test datasets. Figure \ref{convergence_1} shows that the learning algorithm helps improve the makespace reduction for SJF and CP when the training iteration starts. It displays the trend of convergence of our learning algorithm on the test datasets for both of the two algorithms, SJF and CP. Within 10,000 iterations, our learning-based method reaches convergence for both SJF and CP. Also from the results it shows that the convergence is more obvious for the testset with larger number of DAGs (e.g. 15 and 20) compared to the testsets with fewer DAGs (e.g. 5 and 10).

\section{Conclusions}
This paper proposes a novel learning-based approach to improve DAG scheduling by learning to add directional edges in DAGs. We introduce a deep reinforcement learning algorithm on top of a GNN model of DAG structure. Our approach can be easily applied to existing heuristic job scheduling algorithms such as Shortest Job First (SJF), Critical Path (CP), and so on. The experimental results show that it can efficiently improve some heuristic scheduling algorithms for DAG scheduling. As for future work, we plan to extend the support of multi-dimensional resource requirements of DAGs. We will extend the support of other heuristic algorithms such as Tetris and others. Furthermore, it is very interesting to study DAG scheduling under online streaming settings where DAG jobs arrive dynamically over time.



\bibliography{main}

\begin{thebibliography}{10}

\bibitem[Robert Grandl, Ganesh Ananthanarayanan, Srikanth Kandula, Sriram Rao, and
  Aditya Akella(2014)]{grandl2014multi}
Robert Grandl, Ganesh Ananthanarayanan, Srikanth Kandula, Sriram Rao, and
  Aditya Akella.
\newblock Multi-resource packing for cluster schedulers.
\newblock {\em ACM SIGCOMM Computer Communication Review}, 44(4):455--466,
  2014.

\bibitem[Hongzi Mao, Malte Schwarzkopf, Shaileshh~Bojja Venkatakrishnan, Zili Meng, and
  Mohammad Alizadeh(2019)]{mao2019learning}
Hongzi Mao, Malte Schwarzkopf, Shaileshh~Bojja Venkatakrishnan, Zili Meng, and
  Mohammad Alizadeh.
\newblock Learning scheduling algorithms for data processing clusters.
\newblock In {\em Proceedings of the ACM Special Interest Group on Data
  Communication}, pages 270--288. ACM, 2019.

\bibitem[Craig Chambers, Ashish Raniwala, Frances Perry, Stephen Adams, Robert~R. Henry,
  Robert Bradshaw, and Nathan Weizenbaum(2010)]{cham2010}
Craig Chambers, Ashish Raniwala, Frances Perry, Stephen Adams, Robert~R. Henry,
  Robert Bradshaw, and Nathan Weizenbaum.
\newblock Flumejava: Easy, efficient data-parallel pipelines.
\newblock {\em ACM SIGPLAN Conference on Programming Language Design and
  Implementation (PLDI)}, 2010.

\bibitem[Michael Isard, Mihai Budiu, Yuan Yu, Andrew Birrell, and Fetterly Dennis(2007)]{isard2007}
Michael Isard, Mihai Budiu, Yuan Yu, Andrew Birrell, and Fetterly Dennis.
\newblock Dryad: Distributed data-parallel programs from sequential building
  blocks.
\newblock {\em ACM SIGOPS/EuroSys European Conference on Computer Systems
  (EuroSys)}, 2007.

\bibitem[atei Zaharia, Mosharaf Chowdhury, Tathagata Das, Ankur Dave, Justin Ma, Murphy
  McCauley, Michael~J. Franklin, Scott Shenker, and Ion Stoica(2012)]{zaha2012}
Matei Zaharia, Mosharaf Chowdhury, Tathagata Das, Ankur Dave, Justin Ma, Murphy
  McCauley, Michael~J. Franklin, Scott Shenker, and Ion Stoica.
\newblock Dryad: Distributed data-parallel programs from sequential building
  blocks.
\newblock {\em The 9th USENIX Conference on Networked Systems Design and
  Implementation (NSDI)}, 2012.

\bibitem[Mikael Henaff, Joan Bruna, and Yann LeCun(2015)]{lecun2015}
Mikael Henaff, Joan Bruna, and Yann LeCun.
\newblock Deep convolutional networks on graph-structured data.
\newblock {\em CoRR abs/1506.05163}, 2015.

\bibitem[Vijay~V Vazirani(2013)]{vazirani2013approximation}
Vijay~V Vazirani.
\newblock {\em Approximation algorithms}.
\newblock Springer Science \& Business Media, 2013.

\bibitem[Eva Hopper and Brian~CH Turton(2001)]{hopper2001review}
Eva Hopper and Brian~CH Turton.
\newblock A review of the application of meta-heuristic algorithms to 2d strip
  packing problems.
\newblock {\em Artificial Intelligence Review}, 16(4):257--300, 2001.

\bibitem[Silvano Martello, David Pisinger, and Daniele Vigo(2000)]{martello2000three}
Silvano Martello, David Pisinger, and Daniele Vigo.
\newblock The three-dimensional bin packing problem.
\newblock {\em Operations research}, 48(2):256--267, 2000.

\bibitem[Jens Clausen(1999)]{clausen1999branch}
Jens Clausen.
\newblock Branch and bound algorithms-principles and examples.
\newblock {\em Department of Computer Science, University of Copenhagen}, pages
  1--30, 1999.

\bibitem[alter~H Kohler and Kenneth Steiglitz(1974)]{kohler1974characterization}
Walter~H Kohler and Kenneth Steiglitz.
\newblock Characterization and theoretical comparison of branch-and-bound
  algorithms for permutation problems.
\newblock {\em Journal of the ACM}, 21(1):140--156, 1974.

\bibitem[Yu-Kwong Kwok and Ishfaq Ahmad(1999)]{kwok1999static}
Yu-Kwong Kwok and Ishfaq Ahmad.
\newblock Static scheduling algorithms for allocating directed task graphs to
  multiprocessors.
\newblock {\em ACM Computing Surveys (CSUR)}, 31(4):406--471, 1999.

\bibitem[Michael~A Palis, Jing-Chiou Liou, Sanguthevar Rajasekaran, SUNIL SHENDE, and
  David~SL Wei(1995)]{palis1995online}
Michael~A Palis, Jing-Chiou Liou, Sanguthevar Rajasekaran, SUNIL SHENDE, and
  David~SL Wei.
\newblock Online scheduling of dynamic trees.
\newblock {\em Parallel Processing Letters}, 5(04):635--646, 1995.

\bibitem[Guan-Ing Chen and Ten-Hwang Lai(1988)]{chen1988preemptive}
Guan-Ing Chen and Ten-Hwang Lai.
\newblock Preemptive scheduling of independent jobs on a hypercube.
\newblock {\em Information Processing Letters}, 28(4):201--206, 1988.

\bibitem[Hesham Elrewini and Hesham~H Ali(1995)]{elrewini1995static}
Hesham Elrewini and Hesham~H Ali.
\newblock Static scheduling of conditional branches in parallel programs.
\newblock {\em Journal of parallel and Distributed computing}, 24(1):41--54,
  1995.

\bibitem[Hongzi Mao, Mohammad Alizadeh, Ishai Menache, and Srikanth Kandula(2016)]{deeprm}
Hongzi Mao, Mohammad Alizadeh, Ishai Menache, and Srikanth Kandula.
\newblock Resource management with deep reinforcement learning.
\newblock {\em ACM Workshop on Hot Topics in Networks}, 2016.

\bibitem[Xinyun Chen and Yuandong Tian(2019)]{chen2019}
Xinyun Chen and Yuandong Tian.
\newblock Learning to perform local rewriting for combinatorial optimization.
\newblock {\em Advances in Neural Information Processing Systems (NeurIPS)},
  2019.

\bibitem[Kaiming He, Xiangyu Zhang, Shaoqing Ren, and Jian Sun(2016)]{he2016}
Kaiming He, Xiangyu Zhang, Shaoqing Ren, and Jian Sun.
\newblock Deep residual learning for image recognition.
\newblock {\em 29th IEEE Conference on Computer Vision and Pattern Recognition
  (CVPR)}, 2016.

\bibitem[Richard~S. Sutton and Andrew~G. Barto(2018)]{sutton2018}
Richard~S. Sutton and Andrew~G. Barto.
\newblock Reinforcement learning: An introduction.
\newblock {\em Second Edition}, 2018.

\end{thebibliography}

\appendix{}

\section{Algorithms} \label{appalgo}
Here contains the algorithms for our method.

\subsection{Training Algorithm} \label{trainalgo}
The training algorithm of our learning method is given in Algorithm 1.
\begin{algorithm} \label{alg1}
\caption{The training algorithm for learning-based DAG scheduling}
\begin{algorithmic}[1]
\STATE{Function \textbf{Learn}{($\alpha, \phi, \theta_1, \theta_2, I, N, M, H$)}}
\STATE{\# $\alpha$: learning rate, $\phi$: GNN parameters, \# $\theta_1$: policy1 parameters, $\theta_2$: policy2 parameters}
\STATE{\# $I$: num of iterations, $N$: num of rollouts, $M$: num of edges to add, $H$: num of GNN hops}
      \FOR{$i\in\{1,...,I\}$}
           \STATE{$num\_dags$ = Random sample a number from an exponential distribution}
           \STATE{$dags$ = Random sample $num\_dags$ DAGs from a dataset}
           \STATE{$triplets = \textbf{rollout}(dags, N, M, H, \phi, \theta_1, \theta_2)$}
           \STATE{$d_{\phi} = d_{\theta_1} = d_{\theta_2}=0$}
           \FOR{{$(G', (a^1, a^2), \widehat r)\in triplets$}}
               \STATE{$\nabla J(\phi,\theta_1,\theta_2|G') = \widehat r \cdot [\nabla \log p_{\phi,\theta_1}(a^1|G') + \nabla \log p_{\phi,\theta_2}(a^2|G',a^1)]$}   
               \STATE{$d_{\phi} \mathrel{+}= \nabla _{\phi} J(\phi,\theta_1,\theta_2|G')$}
               \STATE{$d_{\theta_1} \mathrel{+}= \nabla _{\theta_1} J(\phi,\theta_1,\theta_2|G')$}
               \STATE{$d_{\theta_2} \mathrel{+}= \nabla _{\theta_2} J(\phi,\theta_1,\theta_2|G')$}
           \ENDFOR
           \STATE{$\phi = \phi + \alpha \cdot d_{\phi}$}
           \STATE{$\theta_1 = \theta_1 + \alpha \cdot d_{\theta_1}$}
           \STATE{$\theta_2 = \theta_2 + \alpha \cdot d_{\theta_2}$}
      \ENDFOR
\end{algorithmic}
\end{algorithm}

\subsection{Rollout Algorithm} \label{rollout}
The rollout algorithm of our learning method is listed in Algorithm 2.

\begin{algorithm} \label{alg2}
\caption{The rollout algorithm for collecting actions and rewards in a DAG graph}
\begin{algorithmic}[1]
\STATE{Function \textbf{Rollout}{($dags, N, M, H, \phi, \theta_1, \theta_2$)}}
      \STATE{\# $dags$: DAG jobs, $N$: num of rollouts, $M$: num of edges to add, $H$: num of GNN hops}
      \STATE{\# $\phi$: GNN parameters, $\theta_1$: parameters for first policy, $\theta_2$: parameters for second policy}
      \STATE{$G$ = Merge $dags$ into one DAG graph}
      \STATE{$T_0 = MakeSpan(G)$} \COMMENT{\#$MakeSpan$: calculated with heuristic scheduling algorithm}
      \STATE{$triplets = [[\ ]] \times N$}
      \FOR{{$i\in\{1,...,N\}$}}
          \STATE{$G_1 = G.copy()$}
          \FOR{{$j\in\{1,...,M\}$}\COMMENT{\# add $M$ edges}}
              \STATE{$graph\_em, node\_ems = GNN(\phi, G_1, H)$} \COMMENT{\#Compute graph  embeddings}
              \STATE{\# Compute starting node}
              \STATE{$qual\_nodes1$ = Identify qualified non-conflict starting nodes in $G$}
              \STATE{$qual\_node1\_ems$ = Lookup embeddings of $qual\_nodes1$ from $node\_ems$}
              \STATE{$qual\_node1\_probs$ = resnet\_policy1($\theta_1, graph\_em, qual\_node1\_ems$)}
              \STATE{$node1$ = Sample a node from $qual\_node1\_probs$ with $\epsilon$-random strategy}
              \STATE{{\# Compute ending node}}
              \STATE{$node1\_em$ = Lookup embedding of $node1$ from $node\_ems$}
              \STATE{$qual\_nodes2$ = Identify qualified non-conflict ending nodes directed by $node1$}
              \STATE{$qual\_node2\_ems$ = Lookup embeddings of $qual\_nodes2$ from $node\_ems$}
              \STATE{$qual\_node2\_probs$ = resnet\_policy2($\theta_2, graph\_em, node1\_em, qual\_node2\_ems$)}
              \STATE{$node2$ = Sample a node from $qual\_node2\_probs$ with $\epsilon$-random strategy}
              \STATE{$G_2$ = Add a directed edge ($node1$, $node2$) in $G_1$}
              \STATE{$record = (G_1, (node1, node2), Makespan(G_1) - Makespan(G_2))$}
              \STATE{$triplets[i].add(record)$}
              \STATE{$G_1 = G_2$}
         \ENDFOR
    \ENDFOR
    \RETURN{\textbf{adjustReward}($triplets, N, M, T_0$)}
\end{algorithmic}
\end{algorithm}

\subsection{Reward Adjustment Algorithm} \label{adjust_reward}
The reward adjustment algorithm of our learning method is listed in Algorithm 3.

\begin{algorithm} \label{alg3}
\caption{The algorithm for adjustment of rewards}
\begin{algorithmic}[1]
\STATE{Function \textbf{adjustReward}{($triplets, N, M, T$)}}
    \STATE{\# adjust and normalize rewards on $N$ rollouts and $M$ added edges}
    \STATE{$baseline = [0.0] \times M$}
    \FOR{$i\in\{1,...,N\}$}
        \FOR{$j\in\{1,...,M\}$} 
             \FOR{$k\in\{j+1,...,M\}$} 
                 \STATE{$triplets[i][j].reward \mathrel{+}= triplets[i][k].reward$}
             \ENDFOR
             \STATE{$baseline[j] \mathrel{+}= (triplets[i][j].reward \ / \ N$)}
         \ENDFOR
    \ENDFOR
    \FOR{$i\in\{1,...,N\}$}
        \FOR{$j\in\{1,...,M\}$}
             \STATE{$triplets[i][j].reward = (triplets[i][j].reward - baseline[j]) \ /\ T$}
         \ENDFOR
    \ENDFOR
    \RETURN{$triplets$}
\end{algorithmic}
\end{algorithm}

\subsection{Inference Algorithm} \label{inf_alg}
The inference algorithm of our learning method is listed in Algorithm 4.

\begin{algorithm} \label{alg4}
\caption{The inference algorithm of adding edges in a DAG}
\begin{algorithmic}[1]
\STATE{Function \textbf{Inference}{($G, M, K, \phi, \theta_1, \theta_2$)}}
    \STATE{\# $G$: graph, $M$: num of edges to add, $K$: num of top candidates}\\
    \STATE{\# $\phi$: GNN params, $\theta_1$: starting node policy params, $\theta_2$: ending node policy params}\\
    \FOR{$i\in\{1,...,M\}$}
        \STATE{$graph\_em, node\_ems = GNN(\phi, G, H)$} \COMMENT{\#Compute graph embeddings}    
        \STATE{$qual\_node1\_ems$ = Get embeddings from $node\_ems$ for qualified starting nodes}              
        \STATE{$qual\_node1\_probs$ = resnet\_policy1($\theta_1, graph\_em, qual\_node1\_ems$)}
        \STATE{$starting\_nodes$ = Pick top K nodes from $qual\_node1\_probs$}
        \STATE{$candidate\_edges = [\ ]$}
        \FOR{$(node1, prob1)\in{starting\_nodes}$}
             \STATE{$qual\_node2\_ems$ = Get embeddings from $node\_ems$ for qualified nodes started by $node1$}
             \STATE{$qual\_node2\_probs$ = resnet\_policy2($\theta_2, graph\_em, node1\_em, qual\_node2\_ems$)}
             \STATE{$node2$, $prob2$ = pick the node with max value in $qual\_node2\_probs$}
             \STATE{$candidate\_edges.add((node1, node2), prob1 \times prob2)$}
        \ENDFOR
        \STATE{$edge$ = Pick ($node1, node2$) from $candidate\_edges$ having max value of $prob$}
        \STATE{$G$ = Add $edge$ in $G$}
    \ENDFOR
    \RETURN{$G$}
\end{algorithmic}
\end{algorithm}

\section{MILP Formulation} \label{milp_form}
Here we define a general MILP formulation for DAG scheduling, where there are N task nodes in a DAG, and the dimension of the resource requirement is $K (K \geq 1)$.

\subsection{Notations}
Below are the coefficients for the MILP problem:
\begin{itemize}
\item $t_i$: duration of task i, total number of tasks is N,
\item $r_i^k$: task i consumption on resource k, dimension of resource is K,
\item $R^k$: total amount of resource in dimension k
\end{itemize}

Below are the decision Variables for the MILP problem:
\begin{itemize}
\item $s_i$: start time of task i
\end{itemize}

Below are some predefined operators for the MILP problem:
\begin{itemize}
\item $A(i)$: all ancestors of task i,
\item $D(i)$: all descendants of task i,
\item $C(i)$: all children of task i
\end{itemize}

\subsection{Primal formulation}  \label{pf}
With task start time $s_i$ as decision variables, two conditions are sufficient for formulating the DAG scheduling problem: 
\begin{itemize}
\item (1) the staring time of each task is later than then ending time of all its parent nodes, and 
\item (2) at the start time of each task, the sum of resource consumption of all running tasks is below the maximum resource capacity. 
\end{itemize}

As a result, the DAG scheduling problem can be formulated as $minmax$ optimization as follows:

\begin{equation}
\begin{array}{llll}
& \min {\max_{i}} (s_i + t_i) & \\
s.t. & s_i + t_i \le s_j & \forall i, j \in C(i)\\
     & \sum_{j=1}^N{\mathbb{I}(s_j \leq s_i < s_j + t_j) \cdot r_j^k} <= R^k & \forall i, k \\
\end{array}
\end{equation}

$\mathbb{I}(\cdot)$ in the above formula is an indicator function with value of 1 if True and 0 else.

\subsection{Reformulation}
The $minmax$ optimization with indicator function is intractable, and we use some intermediate decision variables to reformulate the primal optimization problem.

Intermediate decision variables:
\begin{itemize}
\item T: finishing time of all tasks,
\item $y_{ij} \in \{0, 1\}$: start time of task i is later than start time of task j,
\item $z_{ij} \in \{0, 1\}$: start time of task i is earlier than end time of task j,
\item $u_{ij} \in \{0, 1\}$: task j is running when task i starts
\end{itemize}

Constant:
\begin{itemize}
\item $B$: a sufficiently large positive real number
\end{itemize}

The reformulation of the primal optimization problem is as follows:

\begin{equation}
\begin{array}{llll}
& \min T & & \\
s.t. & s_i + t_i \le T & \forall i & (a) \\
 & s_i + t_i \le s_j & \forall i, j\in C(i) & (b) \\
& s_i \le s_j + By_{ij} & \forall i,j\in \neg(D(i) \cup A(j)) & (c) \\
& s_j \le s_i + B(1-y_{ij}) & \forall i,j\in \neg(D(i) \cup A(j)) & (d)\\
& s_j + t_j \le s_i + Bz_{ij} & \forall i,j\in \neg(D(i) \cup A(j)) & (e)\\
&s_i \le s_j + t_j + B(1-z_{ij}) & \forall i,j\in \neg(D(i) \cup A(j)) & (f)\\
& y_{ij} + z_{ij} - 1\le Bu_{ij} & \forall i,j\in \neg
(D(i)\cup A(i)) & (g)\\
& 2-y_{ij} - z_{ij} \le B(1-u_{ij}) & \forall i,j\in \neg
(D(i)\cup A(i)) & (h)\\
& \sum_{j}u_{ij}r_j^k + r_i^k \le R^k & \forall i, k & (i)\\
& y_{ij}, z_{ij}, u_{ij} \in \{0, 1\} & \forall i,j & (j)
\end{array}
\end{equation}

In the above formulas, (a)-(b) defines the objective to be the total running time; (c) ensures condition $1$ of Primal formulation; (d)-(e) follows the definition of $y_{ij}$; (f)-(g) follows the definition of $z_{ij}$; (h)-(i) follows the definition of $u_{ij}$; and (j) ensures condition $2$ of Primal formulation. 

In this way, the MILP problem can be formulated and solved with existing solvers. However, this MILP problem for DAG scheduling has a large set of integer variables for even a DAG with tens of nodes, which could not be solved by existing solvers. Here we solve the MILP problem by relaxing integer variables to a corresponding LP problem, and we reserve the order of $s_i$ from the LP solution to execute the DAG given a resource limit.
\end{document}